\documentclass[10pt,twocolumn,letterpaper]{article}

\usepackage{amsmath,amsfonts,bm}

\def\eqref#1{equation~\ref{#1}}

\def\1{\bm{1}}

\DeclareMathAlphabet{\mathsfit}{\encodingdefault}{\sfdefault}{m}{sl}
\SetMathAlphabet{\mathsfit}{bold}{\encodingdefault}{\sfdefault}{bx}{n}

\DeclareMathOperator*{\argmax}{arg\,max}
\DeclareMathOperator*{\argmin}{arg\,min}

\usepackage{wacv}
\usepackage{times}
\usepackage{epsfig}
\usepackage{graphicx}        
\usepackage{amsmath}
\usepackage{amssymb}
\usepackage{color}
\usepackage{tabularx}
\usepackage{booktabs}
\usepackage{multirow}
\usepackage{algorithm}
\usepackage{algorithmicx}
\usepackage{algpseudocode}  
\usepackage{array}
\usepackage{xspace}
\usepackage{balance}
\usepackage{enumitem}
\usepackage{mathtools}
\usepackage{rotating}
\usepackage{url}
\usepackage{tabulary}
\usepackage[table]{xcolor}
\usepackage{subcaption}
\usepackage[export]{adjustbox}
\usepackage{ctable}
\usepackage{diagbox}
\usepackage{subcaption}
\usepackage{textcomp}

\usepackage{authblk}

\graphicspath{{./figs/}}

\usepackage[pagebackref=true,breaklinks=true,letterpaper=true,colorlinks,bookmarks=false]{hyperref}

\wacvfinalcopy 

\def\wacvPaperID{101} 

\ifwacvfinal\fi
\begin{document}
\title{GradMix: Multi-source Transfer across Domains and Tasks}

\author[1]{Junnan~Li\thanks{equal contribution}}
\newcommand\CoAuthorMark{\footnotemark[\arabic{footnote}]}
\author[1]{Ziwei~Xu\protect\CoAuthorMark}
\author[1]{Yongkang~Wang}
\author[2]{Qi~Zhao}
\author[1]{Mohan~S.~Kankanhalli}
\affil[1]{School of Computing, National University of Singapore}
\affil[2]{Department of Computer Science and Engineering, University of Minnesota}
\affil[ ]{\tt\small \{lijunnan,ziwei.xu\}@u.nus.edu, yongkang.wong@nus.edu.sg, qzhao@cs.umn.edu, mohan@comp.nus.edu.sg}

\maketitle
\ifwacvfinal\thispagestyle{empty}\fi

\maketitle
\begin{abstract}

The computer vision community is witnessing an unprecedented rate of new tasks being proposed and addressed, 
thanks to the deep convolutional networks' capability to find complex mappings from $\mathcal{X}$ to $\mathcal{Y}$.
The advent of each task often accompanies the release of a large-scale annotated dataset,
for supervised training of deep network.
However, it is expensive and time-consuming to manually label sufficient amount of training data.
Therefore, it is important to develop algorithms that can leverage off-the-shelf labeled dataset to learn useful knowledge for the target task.
While previous works mostly focus on transfer learning from a single source,
we study multi-source transfer across domains and tasks (MS-DTT),
in a semi-supervised setting.
We propose GradMix, 
a model-agnostic method applicable to any model trained with gradient-based learning rule,
to transfer knowledge via gradient descent by weighting and mixing the gradients 
from all sources during training.
GradMix follows a meta-learning objective, 
which assigns layer-wise weights to the source gradients,
such that the combined gradient follows the direction that minimize the loss for a small set of samples from the target dataset.
In addition, we propose to adaptively adjust the learning rate for each mini-batch based on its importance to the target task,
and a pseudo-labeling method to leverage the unlabeled samples in the target domain.
We conduct MS-DTT experiments on two tasks: digit recognition and action recognition,
and demonstrate the advantageous performance of the proposed method against multiple baselines.

\end{abstract} 
\section{Introduction}
\label{sec:introduction}
\begin{figure}[!t]
  \centering
  \includegraphics[width=\linewidth]{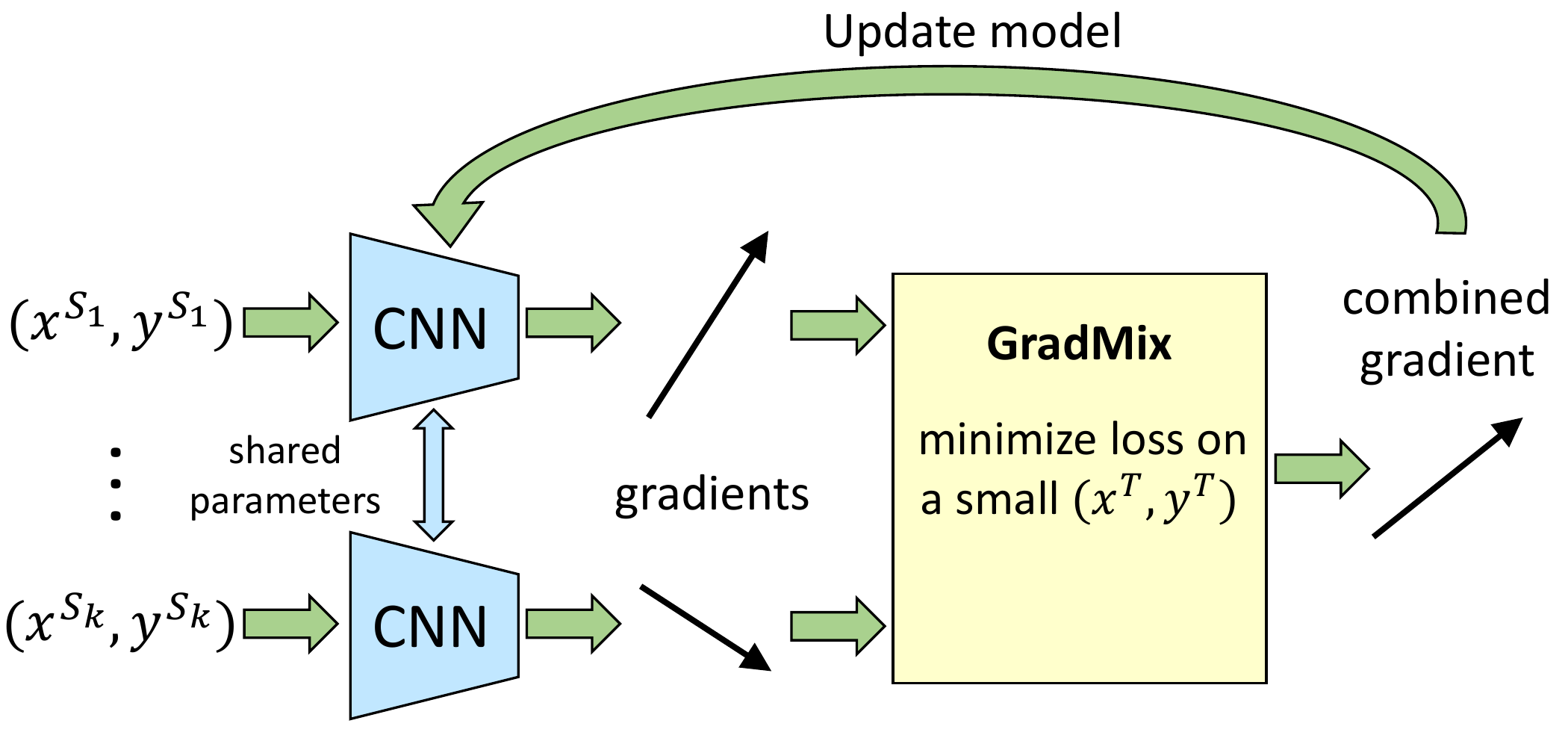}
  	\vspace{-2ex}
  \caption
{
High-level overview of the proposed method. We transfer knowledge to the target domain by weighting and mixing gradients from source domains,
such that the combined gradient should minimize the loss for a few validation samples from the target domain.
}
\label{fig:framework}
\end{figure}  
Deep convolutional networks (ConvNets) have significantly improved the state-of-the-art for visual recognition,
by finding complex mappings from $\mathcal{X}$ to $\mathcal{Y}$.
Unfortunately, these impressive gains in performance come only when massive amounts of paired labeled data $(x,y)$ s.t. $x\in \mathcal{X}, y \in \mathcal{Y}$ are available for supervised training.
For many application domains, it is often prohibitive to manually label sufficient training data, 
due to the significant amount of human efforts required or the concern of violating individual's privacy.
Hence, there is strong incentive to develop algorithms that can reduce the burden of manual labeling,
typically by leveraging off-the-shelf labeled datasets from other related domains and tasks.

There has been a large amount of efforts in the research community to address adapting deep models across domains~\cite{Ganin_ICML_2015,Long_NIPS_2016,Tzeng_CVPR_2017},
to transfer knowledge across tasks~\cite{Luo_NIPS_2017,He_ICCV_2017,Zamir_2018_CVPR},
and to learn efficiently in a few shot manner~\cite{Finn_MAML_2017,Ravi_Meta_2017,Ren_Meta_2018}.
However, most works focus on a single-source and single-target scenario.
Recently, 
some works~\cite{Xu_2018_CVPR,Mancini_CVPR_2018,Zhao_NIPS_2018} propose deep approaches for multi-source domain adaptation,
but assume that the source and target domains have shared label space (task).

In many computer vision applications,
there often exist multiple labeled datasets available from different domains and/or tasks related to the target application.
Hence, it is important and practically valuable that we can transfer knowledge from as many source datasets as possible.
In this work, we formalize this problem as multi-source domain and task transfer (MS-DTT).
Given a set of labeled source dataset,
{\small $\mathcal{S} = \{S_1,S_2,...,S_k\}$},
we aim to transfer knowledge to a sparsely labeled target dataset {\small $\mathcal{T}$}.
Each source dataset {\small $S_i$} could come from a different domain compared to {\small $\mathcal{T}$},
having a different task,
or different in both domain and task.
We focus on a semi-supervised setting where only few samples in {\small $\mathcal{T}$} have labels.

Most works achieve domain transfer by aligning the feature distribution of source domain and target domain~\cite{Long_ICML_2015,Long_NIPS_2016,Ganin_ICML_2015,Tzeng_ICCV_2015,Mancini_CVPR_2018,Xu_2018_CVPR}.
However, this method could be suboptimal for MS-DTT.
The reason is that in MS-DTT, the distribution of source data {\small $p(x^{S_i},y^{S_i})$} and target data {\small $p(x^\mathcal{T},y^\mathcal{T})$} could be significantly
different in both input space and label space, thus feature alignment may generate indiscriminative features for the target classes.
In addition,
feature alignment introduces additional layers and loss terms, which require careful design to perform well.

In this work, we propose a generic and scalable method,
namely GradMix,
for semi-supervised MS-DTT.
GradMix is a model-agnostic method, applicable to any model that uses gradient-based learning rule. 
Our method does not introduce extra layers or loss functions for feature alignment.
Instead, we perform knowledge transfer via gradient descent, 
by weighting and mixing the gradients from all the source datasets during training.
We follow a meta-learning paradigm and model the most basic assumption:
\textit{the combined gradient should minimize the loss for a set of unbiased samples from the target dataset}~\cite{Ren_ICML_2018}.
We propose an online method to weight and mix the source gradients at each training iteration,
such that the knowledge most useful for the target task is preserved through the gradient update.
Our method can adaptively adjust the learning rate for each mini-batch based on its importance to the target task.
In addition, we propose a pseudo-labeling method based on model ensemble to learn from the unlabeled data in target domain.
We perform extensive experiments on two sets of MS-DTT task,
including digit recognition and action recognition,
and demonstrate the advantageous performance of the proposed method compared to multiple baselines. 

\section{Related Work}
\label{sec:literature}

\subsection{Domain Adaptation}
Domain adaptation seeks to address the domain shift problem~\cite{Gabriela_survey_2017} and learn from source domain a model that performs well on the target domain.
Most existing works focus on aligning the feature distribution of the source domain and the target domain.
Several works attempt to learn domain-invariant features by minimizing Maximum Mean Discrepancy~\cite{Long_ICML_2015,Long_NIPS_2016,Sun_ECCVW_2016}.
Other methods propose adversarial discriminative models,
which try to learn domain-agnostic representations by maximizing a domain confusion loss~\cite{Ganin_ICML_2015,Tzeng_ICCV_2015,Luo_NIPS_2017}.

Recently,
multi-source domain adaptation with deep model has been studied.
Mancini~\etal~\cite{Mancini_CVPR_2018} use DA-layers~\cite{Fabio_ICCV_2017,Li_arxiv_2016} to minimize the distribution discrepancy of network activations.
Xu~\etal~\cite{Xu_2018_CVPR} propose multi-way adversarial domain discriminator that minimizes the domain discrepancies between the target and each of the sources.
Zhao~\etal~\cite{Zhao_NIPS_2018} propose multisource domain adversarial networks that
approach domain adaptation by optimizing domain-adaptive generalization bounds.
However, all of these methods~\cite{Mancini_CVPR_2018,Xu_2018_CVPR,Zhao_NIPS_2018} assume that the source and target domains have a shared label space.

\subsection{Transfer Learning.}
Transfer learning extends domain adaptation into more general cases,
where the source and target domain could be different,
in both input space and label space~\cite{Sinno_survey_2010,Weiss_survey_2016,Junnan_NIPS,Junnan_MM}.
In computer vision,
transfer learning has been widely studied to overcome the deficit of labeled data by adapting models trained for other tasks.
With the advance of deep supervised learning,
ConvNets trained on large datasets such as ImageNet~\cite{Russakovsk_IJCV_2015} have achieved state-of-the-art performance when transfered to other tasks
(\eg~object detection~\cite{He_ICCV_2017}, semantic segmentation~\cite{Long_CVPR_2015}, etc.)
by simple fine-tuning.
In this work, we focus on the setting where source and target domains have the same input space and different label spaces.

\subsection{Meta-Learning.}
Meta-learning aims to utilize knowledge from past experiences to learn quickly on target tasks, from only a few annotated samples.
Meta-learning generally seeks performing the learning at a level higher than where conventional learning occurs,
\eg~learning the update rule of a learner~\cite{Ravi_Meta_2017},
or finding a good initialization point that is more robust~\cite{MLNT} or can be easily fine-tuned~\cite{Finn_MAML_2017}.
Li~\etal~\cite{Li_AIII_2018} propose a meta-learning method to train models with good generalization ability to novel domains.
Franceschi~\etal~\cite{Franceschi_2018_ICML} introduce a framework based on bilevel programming that unifies gradient-based hyperparameter optimization and meta-learning.
Sun~\etal~\cite{Sun_2019_CVPR} propose a meta-transfer learning method to address the few-shot learning task.
Ren~\etal~\cite{Ren_ICML_2018} propose example reweighting in a meta-learning framework.
Our method follows the meta-learning paradigm that uses validation loss as the meta-objective.
However, different from~\cite{Ren_ICML_2018} which reweight samples in a batch for robust learning against noise,
we reweight source domain gradients layer-wise for transfer learning.
Gradient alignment has also been used to enhance learning congruency in~\cite{Yan_PAMI}.
\section{Method}
\label{sec:method}
\subsection{Problem Formulation}
We first formally introduce the semi-supervised MS-DTT problem.
Assume that there exists a set of $k$ source domains {\small $\mathcal{S} = \{S_1,S_2,...,S_k\}$} and a target domain {\small $\mathcal{T}$}.
Each source domain {\small $S_i$} contains {\small $N^{S_i}$} images,
{\small $x^{S_i} \!\! \in \! \mathcal{X}^{S_i}$},
with associated labels {\small $y^{S_i} \!\! \in \! \mathcal{Y}^{S_i}$}.
Similarly,
the target domain consists of {\small $N^\mathcal{T}$} unlabeled images,
{\small $x^\mathcal{T} \!\! \in \! \mathcal{X}^{\mathcal{T}}$},
as well as {\small $M^\mathcal{T}$} labeled images with associated labels {\small $y^\mathcal{T} \!\! \in \! \mathcal{Y}^{\mathcal{T}}$}.
We assume target domain is only sparsely labeled,
\ie~{\small $M^\mathcal{T} \!\! \ll \! N^\mathcal{T}$}.
Our goal is to learn a strong target classifier that can predict labels {\small $y^\mathcal{T}$} given {\small $x^\mathcal{T}$}.

Different from standard domain adaptation approaches that assume a shared label space between each source and target domain ({\small $\mathcal{Y}^{S_i} \! = \! \mathcal{Y}^\mathcal{T}$}), we study the problem of joint transfer across domains and tasks.
In our setting, only one of the source domain needs to have the same label space as the target domain 
({\small $\exists S_i ~\textit{s}.\textit{t}. ~\mathcal{Y}^{S_i} = \mathcal{Y}^\mathcal{T}$}).
Other source domains could either have a partially overlapping label space with the target domain 
({\small $\mathcal{Y}^{S_i} \cap \mathcal{Y}^\mathcal{T} \subset \mathcal{Y}^\mathcal{T}$} and {\small $\mathcal{Y}^{S_i} \cap \mathcal{Y}^\mathcal{T} \ne \varnothing$}), 
or a non-overlapping label space ({\small $\mathcal{Y}^{S_i} \cap \mathcal{Y}^\mathcal{T} = \varnothing$}).

\subsection{Meta-learning Objective}

Let $\Theta$ denote the network parameters for our model.
We consider a loss function {\small $\mathcal{L}(x,y;\Theta) = f(\Theta)$} to minimize during training.
For deep networks, stochastic gradient descent (SGD) or its variants are commonly used to optimize the loss functions.
At every step $n$ of training,
we forward a mini-batch of samples from each of the source domain {\small $\{S_i\}^k_{i=1}$},
and apply back-propagation to calculate the gradients w.r.t the parameters $\Theta_n$,
{\small $\nabla f_{s_i}(\Theta_n)$}.
The parameters are then adjusted according to the sum of the source gradients.
For example, for vanilla SGD:
\begin{equation}
	\Theta_{n+1}=\Theta_n-\alpha \sum_{i=1}^{k} \nabla f_{s_i}(\Theta_n),
\end{equation}
where $\alpha$ is the learning rate.

In semi-supervised MS-DTT, we also have a small validation set $\mathcal{V}$ that contains few labeled samples from the target domain.
We want to learn a set of weights for the source gradients,
$w = \{w_{s_i}\}^k_{i=1}$,
such that when taking a gradient descent using their weighted combination $\sum_{i=1}^{k} w_{s_i} \nabla f_{s_i}(\Theta_n)$,
the loss on the validation set is minimized:
\begin{align}
	\Theta^\ast(w) &= \Theta_n-\alpha \sum_{i=1}^{k} w_{s_i} \nabla f_{s_i}(\Theta_n), \\
	\label{eqn:3}
	w^\ast &= \argmin_{w,w\ge 0} f_\mathcal{V}(\Theta^\ast(w))
\end{align}

\subsection{Layer-wise Gradient Weighting}
\label{sec:gradmix}
Calculating the optimal $w^\ast$ requires two nested loops of optimization,
which can be computationally expensive.
Here we propose an approximation to the above objective.
At each training iteration $n$,
we do a forward-backward pass using the small validation set $\mathcal{V}$ to calculate the gradient,
{\small $\nabla f_\mathcal{V}(\Theta_n)$}.
We take a first-order approximation and assume that adjusting {\small $\Theta_n$} in the direction of {\small $\nabla f_\mathcal{V}(\Theta_n)$} can minimize {\small $f_\mathcal{V}(\Theta_n)$}.
Therefore, we find the optimal $w^\ast$ by maximizing the cosine similarity between the combined source gradient and the validation gradient:
\begin{equation}
	\! w^\ast \! = \argmax_{w,w\ge 0} \text{cossim}\left[\sum\limits_{i=1}^{k} w_{s_i} \nabla f_{s_i}(\Theta_n), \nabla f_\mathcal{V}(\Theta_n)\right],
	\label{eqn:w}
\end{equation}
where the cosine similarity between two vectors is defined as:
\begin{equation}
	\text{cossim}[a,b] = \frac{a\cdot b}{\left\Vert a \right\Vert \left\Vert b \right\Vert}.
\end{equation}


Instead of using a global weight value for each source gradient,
we propose a layer-wise gradient weighting,
where the gradient for each network layer are weighted separately.
This enables a finer level of gradient combination.
Specifically,
in our MS-DTT setting,
all source domains and the target domain share the same parameters up to the last fully-connected (fc) layer,
which is task-specific (the target domain shares its last layer only with the source domain that has the same label space as the target).
Therefore, for each layer $l$ with parameter $\theta^l$,
and for each source domain $S_i$,
we have a corresponding weight $w_{s_i}^l$.
We can then write Equation~\ref{eqn:w} as:
\begin{equation}
	\label{eqn:w_layer}
	w^\ast = 
	\argmax_{w,w\ge 0}\sum_{l=1}^{L-1} \text{cossim}\left[
	\sum_{i=1}^{k} w_{s_i}^l \nabla f_{s_i}(\theta_n^l),
	\nabla f_\mathcal{V}(\theta_n^l)\right],
\end{equation}

where $L$ is the total number of layers for the ConvNet.
We constrain {\small $w_{s_i}^l\ge0$} for all $i$ and $l$,
since negative gradient update can usually result in unstable behavior.
To efficiently solve the above constrained non-linear optimization problem,
we utilize a sequential quadratic programming method,
SLSQP, implemented in NLopt~\cite{NLopt}.

In practice, we normalize the weights for each layer across all source domains so that they sum up to one:
\begin{equation}
	\tilde{w}_{s_i}^l = \frac{w_{s_i}^l}{\sum_{i=1}^{k}w_{s_i}^l}
\end{equation}

The computational overhead of GradMix mainly comes from optimizing $w$ and calculating {\small $\nabla f_\mathcal{V}$}.
Compared to source-only training,
GradMix increases the training time per-batch by approximately $40\%$.

\subsection{Adaptive Learning Rate}
\label{sec:adalr}

Intuitively,
certain mini-batches from the source domains contain more useful knowledge that can be transferred to the target domain,
whereas some mini-batches contain less.
Therefore, we want to adaptively adjust our training to pay more attention to the important mini-batches.
To this end, we measure the importance score $\rho$ of a mini-batch using the cosine similarity between the optimally combined gradient and the validation gradient:
\begin{equation}
	\rho = \sum_{l=1}^{L-1}
	 \text{cossim}\left[ \sum_{i=1}^{k} \tilde{w}_{s_i}^l \nabla f_{s_i}(\theta_n^l),
	\nabla f_\mathcal{V}(\theta_n^l) \right]
\end{equation}


Based on $\rho$, we calculate a scaling term $\eta$ bounded between 0 and 1:
\begin{equation}
	\label{eqn:lr}
	\eta = \frac{1}{1+e^{-(\beta \rho-\gamma)}},
\end{equation}
where $\beta$ controls the rate of saturation for $\eta$, and $\gamma$ controls the shift along the horizontal axis (\ie~when $\beta \rho=\gamma$, $\eta = 0.5$).
We determine the value of $\beta$ and $\gamma$ empirically through experiments.

Finally, we multiply $\eta$ to the learning rate $\alpha$, and perform SGD to update the parameters:
\begin{equation}
\theta_{n+1}^l=\theta_n^l-\eta \alpha \sum_{i=1}^{k} \tilde{w}_{s_i}^l \nabla  f_{s_i}(\theta_n^l), \text{~for}~ l={1,2,...,L-1}
\end{equation}

\subsection{Pseudo-label with Ensembles}

In our semi-supervised MS-DTT setting,
there also exists a large set of unlabeled images in the target domain,
denoted as {\small $\mathcal{U}=\{(x_n^\mathcal{T})\}_{n=1}^{N^\mathcal{T}}$}.
We want to learn target-discriminative knowledge from {\small $\mathcal{U}$}.
To achieve this, 
we propose a method to calculated pseudo-labels {\small $\hat{y}_n^\mathcal{T}$} for the unlabeled images,
and construct a pseudo-labeled dataset {\small $S_u=\{(x_n^\mathcal{T},\hat{y}_n^\mathcal{T})\}_{n=1}^{N^p}$}. 
Then we leverage {\small $S_u$} using the same gradient mixing method as described above.
Specifically, 
we consider to minimize a loss {\small $\mathcal{L}_u(x,\hat{y};\Theta)$} during training where {\small $(x,\hat{y}) \in S_u$}.
At each training iteration $n$, we sample a mini-batch from {\small $S_u$}, calculate the gradient {\small $\nabla f_{s_u}(\Theta_n)$},
and combine it with the source gradients {\small $\{\nabla f_{s_i}(\Theta_n)\}_{i=1}^k$} using the proposed layer-wise weighting method.

In order to acquire the pseudo-labels,
we perform a first step to train a model using the source domain datasets following the proposed gradient mixing method,
and use the learned model to label {\small $\mathcal{U}$}.
However, the learned model would inevitably create some false pseudo-labels. 
Previous studies found that ensemble of models helps to produce more reliable pseudo-labels~\cite{Saito_Pseudo_2017,Laine_ensemble_2016}.
Therefore, in our first step,
we train multiple models with different combination of $\beta$ and $\gamma$ in Equation~\ref{eqn:lr}.
Then we pick the top $R$ models with the best accuracies on the hyper-validation set (we set $R=3$ in our experiments),
and use their ensemble to create pseudo-labels.
The difference in hyper-parameters during training ensures that different models learn significantly different sets of weight,
hence the ensemble of their prediction is less biased.

Here we propose two approaches to create pseudo-labels, namely hard label and soft label:
%

\textbf{Hard label.}
Here,
we assume that the pseudo-label is more likely to be correct if all the models can reach an agreement with high confidence.
We assign a pseudo-label {\small $\hat{y}=C$} to an image {\small $x\in \mathcal{U}$},
where $C$ is a class index,
if the two following conditions are satisfied.
First, all of the $R$ models should predict $C$ as the class with maximum probability.
Second, for all models, the probability for $C$ should exceed certain threshold,
which is set as 0.8 in our experiments.
If these two conditions are satisfied, 
we will add $(x,\hat{y})$ into $S_u$.
During training, the loss $\mathcal{L}_u(x,\hat{y};\Theta)$ is the standard cross entropy loss.

\textbf{Soft label.}
Let $p_r$ denote the output from the $r$-th model's softmax layer for an input $x$,
which represents the probability over classes.
We calculate the average of $p_r$ across all of the $R$ models as the soft pseudo-label for $x$,
\ie~{\small $\hat{y}=\frac{1}{R}\sum_{r=1}^{R} p_r$}.
Every unlabeled image {\small $x\in \mathcal{U}$} will be assigned a soft label and added to $S_u$.
During training, let $p_\Theta$ be the output probability from the model,
we want to minimize the KL-divergence between $p_\Theta$ and the soft pseudo-label for all pairs {\small $(x,\hat{y}) \in S_u$}.
Therefore, the loss is {\small $\mathcal{L}_u(x,\hat{y};\Theta) = D_{KL}(p_\Theta,\hat{y})$}.

For both hard label and soft label approach,
after getting the pseudo-labels,
we train a model from scratch using all available datasets {\small $\{S_i\}_{i=1}^k$}, {\small $S_u$} and {\small $\mathcal{V}$}.
Since the proposed gradient mixing method relies on {\small $\mathcal{V}$} to estimate the model's performance on the target domain,
we enlarge the size of {\small $\mathcal{V}$} to 100 samples per class, by adding hard-labeled images from {\small $S_u$} using the method described above.
The enlarged {\small $\mathcal{V}$} can represent the target domain with less bias, which helps to calculate better weights on the source gradients,
such that the model's performance on the target domain is maximized.

\subsection{Incorporating Semi-supervised Learning}

We can further exploit the unlabeled target domain data {\small $\mathcal{U}$} by leveraging semi-supervised learning (SSL) methods.
Specifically,
we incorporate two state-of-the-art SSL methods,
virtual adversarial training~\cite{vat} and MixMatch~\cite{mixmatch},
into our GradMix method,
by adding an additional unlabeled loss term on {\small $\mathcal{U}$} during training.
The details of the unlabeled loss can be found in the original papers~\cite{vat,mixmatch}.
 \begin{figure*}[!t]
	\centering
		\begin{minipage}{1.0\textwidth}
			\begin{minipage}{0.11\textwidth}				\vspace{-3ex}				Digit \\ Recognition			\end{minipage}
			\begin{minipage}{0.89\textwidth}				\centerline{\includegraphics[width=\linewidth]{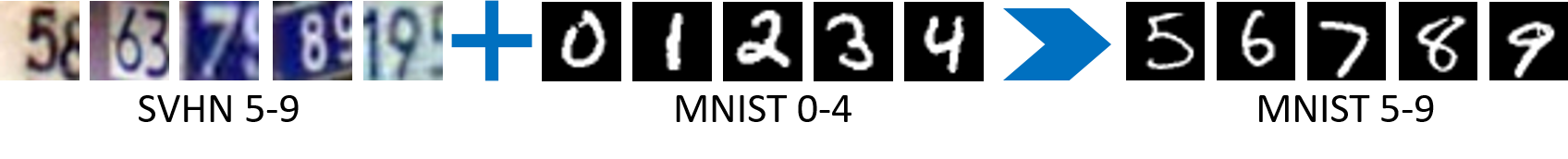}}			\end{minipage}
		\end{minipage}
		\begin{minipage}{1.0\textwidth}
			\begin{minipage}{0.11\textwidth}				\vspace{-2ex}				Action\\ Recognition			\end{minipage}
			\begin{minipage}{0.89\textwidth}				\centerline{\includegraphics[trim={0 17 0 0},clip,width=\linewidth]{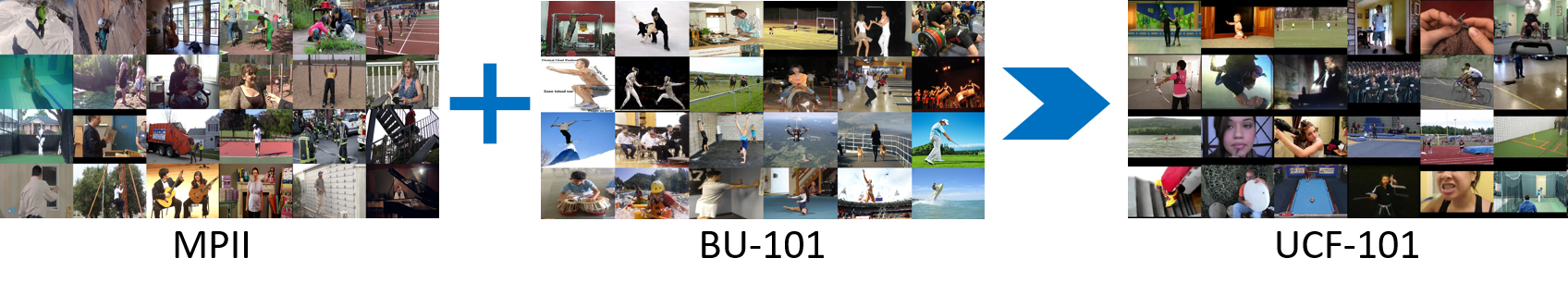}}			\end{minipage}
		\end{minipage}
	\caption
		{
		An illustration of the two experimental settings for multi-source domain and task transfer (MS-DTT).
		Our method effectively transfers knowledge from multiple sources to the target task.
		}
	\label{fig:example}
\end{figure*}

\section{Experiment}
\label{sec:experiment}

\subsection{Experimental Setup}
\noindent\textbf{Datasets.}
In our experiment, 
we perform MS-DTT across two different groups of data settings, as shown in Figure~\ref{fig:example}.
First, we do transfer learning across different digit domains using 
MNIST~\cite{LeCun_MNIST_1998} and Street View House Numbers (SVHN)~\cite{Netzer_NIPSW_2011}.
MNIST is a popular benchmark for handwritten digit recognition,
which contains a training set of 60,000 examples and a test set of 10,000 examples. 
SVHN is a real-word dataset consisting of images with colored background and blurred digits.
It has 73,257 examples for training and 26,032 examples for test.

For our second setup,
we study MS-DTT from human activity images in MPII dataset~\cite{Andriluka_CVPR_2014} 
and human action images from the Web (BU101 dataset)~\cite{Ma_PR_2017}, to video action recognition using UCF101~\cite{Soomro_ucf_2012} dataset.
MPII dataset consists of 28,821 images
covering 410 human activities including home activities, religious activities, occupation, etc.
UCF101 is a benchmark action recognition dataset collected from YouTube.
It has 13,320 videos from 101 action categories,
captured under various lighting conditions with camera motion and occlusion.
We take the first split of UCF101 for our experiment.
BU101 contains 23,800 images collected from the Web,
with the same action categories as UCF101.
It contains professional photos, commercial photos, and artistic photos,
which differ significantly from video frames. 

\noindent\textbf{Network and implementation details.}
For digit recognition, 
we use the same ConvNet architecture as~\cite{Luo_NIPS_2017}, 
which has 4 Conv layers and 2 fc layers. 
We randomly initialize the weights,
and train the network using SGD with learning rate $\alpha=0.05$,
and a momentum of 0.9.
For fine-tuning we reduce the learning rate to 0.005.
For action recognition, 
we use ResNet-18~\cite{He_CVPR_2016} architecture.
We initialize the network with ImageNet pre-trained weights,
which is important for all baseline methods to perform well.
The learning rate is 0.001 for training and $5\mathrm{e}{-5}$ for fine-tuning.
\begin{table*}[!t]
	\centering
	\caption
		{
			Classification accuracy (\%) of the baselines and our method on the test split of MNIST 5-9.
			We report the mean and the standard error of each method across 10 runs with different randomly sampled $\mathcal{V}$.
		}
	\label{tbl:result_digit}

	\begin{tabular}{l|l|c|c|c|c} 
		\toprule	
		\textbf{Method}\hspace{22ex} & \textbf{Datasets}\hspace{4ex} & \textbf{k=2} & \textbf{k=3} & \textbf{k=4} & \textbf{k=5}\\
		\midrule
		Target only & $\mathcal{V}$ &  71.35$\pm$1.85 & 77.15$\pm$1.36&81.43$\pm$1.41 &84.83$\pm$1.10\\
		Source only & $S_1,S_2$ &82.39\hspace{6ex} & 82.39\hspace{6ex} & 82.39\hspace{6ex} & 82.39\hspace{6ex} \\	
			
		Fine-tune & $S_1,S_2,\mathcal{V}$ &89.94$\pm$0.35&89.86$\pm$0.46 &90.89$\pm$0.48 &91.96$\pm$0.39\\
		GradMix SGD~\cite{Ren_ICML_2018} & $S_1,S_2,\mathcal{V}$ &89.30$\pm$0.73& 89.78$\pm$0.72 &91.70$\pm$0.45& 92.05$\pm$0.29 \\
		GradMix w/o AdaLR & $S_1,S_2,\mathcal{V}$ &90.10$\pm$0.37 &90.22$\pm$0.62 &92.14$\pm$0.43 & 92.92$\pm$0.29\\
		GradMix & $S_1,S_2,\mathcal{V}$ & \bfseries{91.17$\pm$0.37} &\bfseries{91.45$\pm$0.52} &\bfseries{92.14$\pm$0.40} &\bfseries{93.06$\pm$0.46}\\	
		\midrule
		MME~\cite{SSL} & $S_1,\mathcal{V},\mathcal{U}$ &90.25$\pm$0.31 &90.37$\pm$0.36 &91.38$\pm$0.29 &91.76$\pm$0.24\\				
		MDDA~\cite{Mancini_CVPR_2018} & $S_1,S_2,\mathcal{V},\mathcal{U}$ &90.23$\pm$0.40 &90.28$\pm$0.50 &91.45$\pm$0.37 &91.85$\pm$0.31\\			
		DCTN~\cite{Xu_2018_CVPR} & $S_1,S_2,\mathcal{V},\mathcal{U}$ &91.81$\pm$0.26 & 92.34$\pm$0.28&92.42$\pm$0.39 &92.97$\pm$0.37\\
		GradMix w/ soft label & $S_1,S_2,\mathcal{V},\mathcal{U}$ &94.62$\pm$0.18 &95.03$\pm$0.30&95.26$\pm$0.17 &95.74$\pm$0.21 \\
		GradMix w/ hard label& $S_1,S_2,\mathcal{V},\mathcal{U}$ &96.02$\pm$0.24 &96.24$\pm$0.33 &96.63$\pm$0.17&96.84$\pm$0.20\\
		GradMix w/ VAT~\cite{vat} & $S_1,S_2,\mathcal{V},\mathcal{U}$ &96.23$\pm$0.21 &96.35$\pm$0.31&\bfseries{96.87}$\pm$\bfseries{0.19} &96.94$\pm$0.20 \\				
		GradMix w/ MixMatch~\cite{mixmatch}& $S_1,S_2,\mathcal{V},\mathcal{U}$ &\bfseries{96.30}$\pm$0.23 &\bfseries{96.43}$\pm$0.32 &96.85$\pm$0.19&\bfseries{97.02}$\pm$\bfseries{0.21}\\
		\bottomrule
	\end{tabular}
	
\end{table*}	
\subsection{SVHN 5-9 + MNIST 0-4 $\rightarrow$ MNIST 5-9}
\label{sec:exp_digit}
\noindent\textbf{Experimental setting.}
In this experiment,
we define four sets of training data:
(1) labeled images of digits 5-9 from the training split of SVHN dataset as the first source $S_1$,
(2) labeled images of digits 0-4 from the training split of MNIST dataset as the second source $S_2$,
(3) few labeled images of digits 5-9 from the training split of MNIST dataset as the validation set $\mathcal{V}$,
(4) unlabeled images from the rest of the training split of MNIST 5-9 as $\mathcal{U}$.
We subsample $k$ examples from each class of MNIST 5-9 to construct the unbiased validation set $\mathcal{V}$.
We experiment with $k=2,3,4,5$,
which corresponds to $10,15,20,25$ labeled examples.
Since $\mathcal{V}$ is randomly sampled, we repeat our experiment 10 times with different $\mathcal{V}$.
In order to monitor training progress and tune hyper-parameters (\eg~ $\alpha,\beta,\gamma$),
we split out another 1000 labeled samples from MNIST 5-9 as the hyper-validation set.
The hyper-validation set is the traditional validation set and is fixed across 10 runs.

\noindent\textbf{Baselines.}
We compare the proposed method to multiple baseline methods:
\begin{itemize}[wide, labelwidth=!, labelindent=0pt]
\item \textit{Target only}: the model is trained using $\mathcal{V}$.
\item \textit{Source only}: the model is trained using $S_1$ and $S_2$ without gradient reweighting.
\item \textit{Fine-tune}: the \textit{Source only} model is fine-tuned using $\mathcal{V}$.
\item \textit{MME}~\cite{SSL}: Minimax Entropy is a state-of-the-art method for single-source semi-supervised domain adaptation.  
We use $S_1$ (SVHN 5-9) as the source domain because it is has the same label space as the target task.
\item \textit{MDDA}~\cite{Mancini_CVPR_2018}: Multi-domain domain alignment layers that shift the network activations for each domain using a parameterized transformation equivalent to batch normalization.  
\item \textit{DCTN}~\cite{Xu_2018_CVPR}: Deep Cocktail Network,
which uses multi-way adversarial adaptation to align the distribution of multiple source domains and the target domain.
\end{itemize}

%

We also evaluate different variants of our model with and without certain component to show its effect:
\begin{itemize}[wide, labelwidth=!, labelindent=0pt]
\item \textit{GradMix SGD}: instead of calculating the optimal weights $w^\ast$ by maximizing cosine similarity of gradients (Equation~\ref{eqn:w_layer}), we follow the method in~\cite{Ren_ICML_2018} and perform SGD on $w$ to directly minimize the validation error in Equation~\ref{eqn:3}. 
\item \textit{GradMix w/o AdaLR}: the method in Section~\ref{sec:gradmix} without the adaptive learning rate (Section~\ref{sec:adalr}).
\item \textit{GradMix}: the proposed method that uses $S_1,S_2$ and $\mathcal{V}$ during training.
\item \textit{GradMix w/ hard label}: using the hard label approach to create pseudo-labels for $\mathcal{U}$, and train a model with all available datasets.
\item \textit{GradMix w/ soft label}: using the soft label approach to create pseudo-labels for $\mathcal{U}$, and train a model with all available datasets.
\item \textit{GradMix w/ VAT}: incorporating VAT~\cite{vat} into GradMix.
\item \textit{GradMix w/ MixMatch}: incorporating MixMatch~\cite{mixmatch} into GradMix.
\end{itemize}

\begin{figure}[!t]
  \centering
%
%
  \begin{minipage}{0.92\columnwidth}  \centerline{\includegraphics[width=\linewidth]{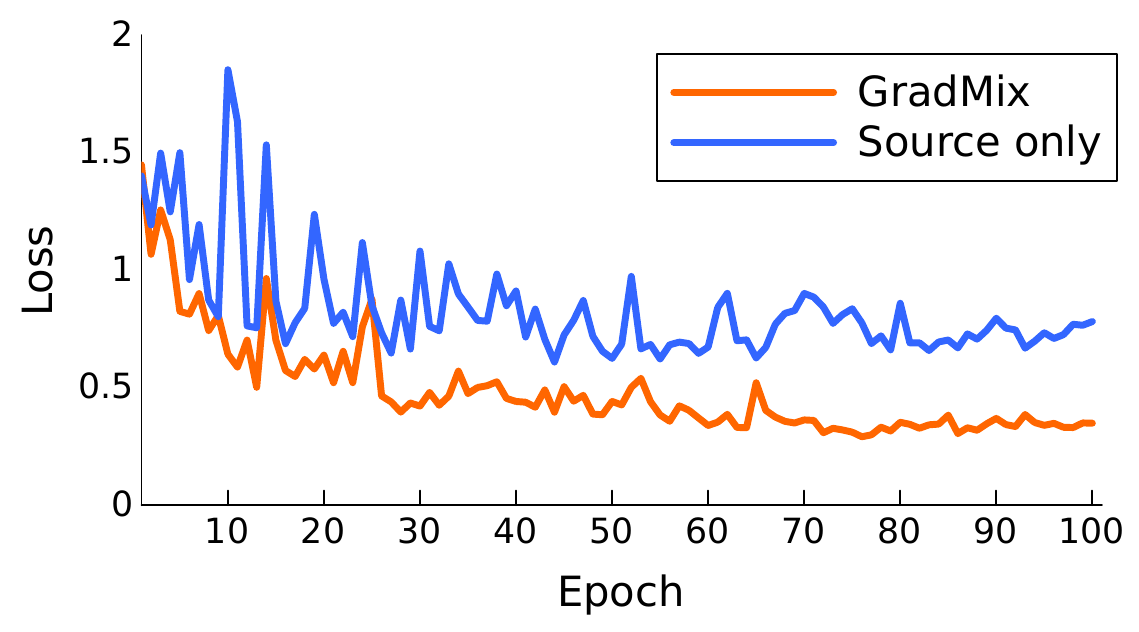}}  \end{minipage}
  \begin{minipage}{0.92\columnwidth}  \centerline{\includegraphics[width=\linewidth]{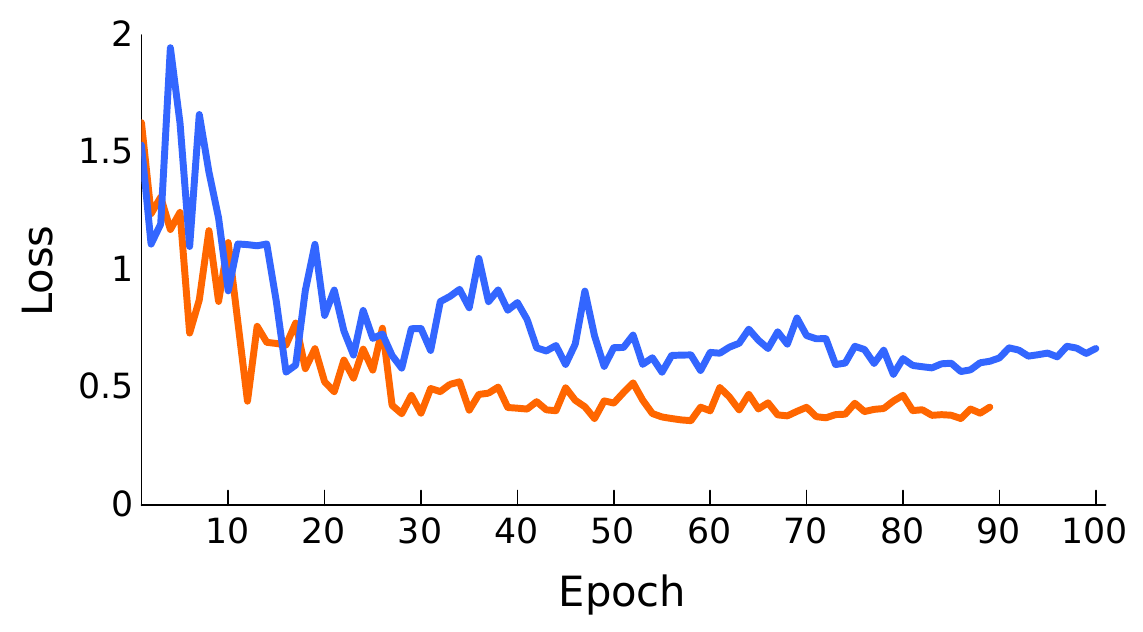}}  \end{minipage}
  \caption
	{
		Loss on the hyper-validation set as training proceeds on digit recognition task. 
		Top row is with $k=2$ whereas the bottom row is with $k=5$.
		We define 1 epoch as training for 100 mini-batches (gradient descents).
	}
\label{fig:loss}
\end{figure}
 \begin{table*}[!t]
	\centering
	\caption
		{
			Results of GradMix using different $\beta$ and $\gamma$ when $k=3$. Numbers indicate the test accuracy (\%) on MNIST 5-9 (averaged across 10 runs).
			The ensemble of the top three models is used to create pseudo-labels.
		}
	\label{tbl:AdaLR}
	\begin{tabular}{l|c|c|c|c|c|c|c|c|c} 
		\toprule	
         &$\gamma=0$&$\gamma=0.1$&$\gamma=0.2$&$\gamma=0.3$&$\gamma=0.4$&$\gamma=0.5$&$\gamma=0.6$&$\gamma=0.7$&$\gamma=0.8$\\
         \midrule
         $\beta = 5$& 90.92 & \textbf{90.96} & 90.95 & 90.58 & 90.75 &90.75 &90.51 & 90.63 & \textbf{91.12}\\
         $\beta = 6$& 90.41 & 90.75 & 89.95 & 90.79 & 90.59 &89.95 &90.58 & 90.63 & 90.56\\      
         $\beta = 7$& 89.76 & 90.44 & 90.42 & 90.94 & 90.28 &90.40 &90.52 & 90.70 & 90.66\\   
         $\beta = 8$& 90.05 & 90.89 & 90.93 & 90.57 & 90.77 &90.69 &89.99 & 90.58 & 90.71\\       
         $\beta = 9$& 90.32 & 90.70 & 90.48 & 90.94 & 90.47 &90.92 &90.20 & 90.23 & 90.86\\   
         $\beta = 10$& 90.52 & 90.03 & 89.67 & 90.01 & 89.84 &90.51 &\textbf{91.45} & 90.58 & 90.70\\                                          
		\bottomrule
\end{tabular}
\end{table*}			
\noindent\textbf{Results.}
Table~\ref{tbl:result_digit} shows the results for methods described above.
We report the mean and standard error of classification accuracy across 10 runs with randomly sampled $\mathcal{V}$.
Methods in the upper part of the table do not use the unlabeled target domain data $\mathcal{U}$.
Among these methods, the proposed GradMix has the best performance.
If we remove the adaptive learning rate,
the accuracy would decrease.
As expected, the performance improves as $k$ increases,
which indicates more samples in $\mathcal{V}$ can help the GradMix method to better combine the gradients during training.

The lower part of the table shows methods that leverage the unlabeled target data $\mathcal{U}$. 
MME~\cite{SSL} only uses $S_1$,
whereas other methods use both $S_1$ and $S_2$.
The proposed GradMix without $\mathcal{U}$ can achieve comparable performance with state-of-the-art baselines that use $\mathcal{U}$ (MME, MDDA and DCTN).
Using pseudo-label with model ensemble significantly improves performance compared to baseline methods.
Comparing soft label to hard label, the hard label approach achieves better performance.
More detailed results about model ensemble for pseudo-labeling is shown later in the ablation study.  
Furthermore,
both VAT~\cite{vat} and MixMatch~\cite{mixmatch} can achieve performance improvement by effectively utilizing the unlabeled data $\mathcal{U}$.

\noindent\textbf{Ablation Study.}
In this section, we perform ablation experiments to demonstrate the effectiveness of our method and the effect of different hyper-parameters.
First, Figure~\ref{fig:loss} shows two examples of the hyper-validation loss as training proceeds.
We show the loss for the \textit{Source only} baseline and the proposed GradMix,
where we perform hyper-validation every 100 mini-batches (gradient descents).
In both examples with different $k$,
GradMix achieves a quicker and steadier decrease in the hyper-validation loss.

In Table~\ref{tbl:AdaLR},
we show the results using GradMix with different combination of $\beta$ and $\gamma$ when $k=3$. 
We perform a grid search with $\beta=[5,6,...,10]$ and $\gamma=[0,0.1,...,0.8]$.
The accuracy is the highest for $\beta=10$ and $\gamma=0.6$.
The top three models are selected for ensemble to create pseudo-labels for the unlabeled set $\mathcal{U}$.

\begin{figure}[!t]
  \centering
  \includegraphics[width=\linewidth]{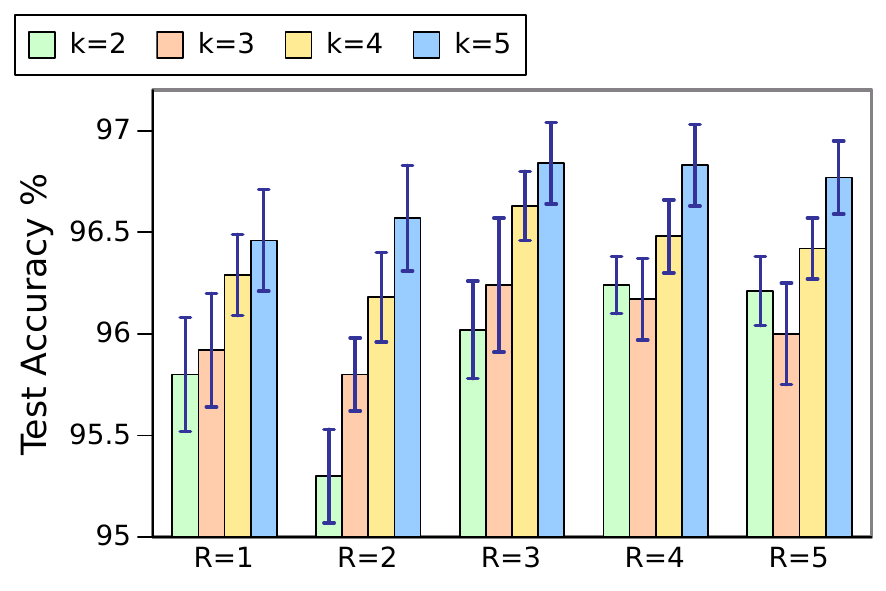}
  	\vspace{-2ex}
  \caption
{
Results of GradMix w/ hard label using various number of pre-trained models (R) for ensemble on digit recognition task. k is the number of labeled samples per class in $\mathcal{V}$.
}
\label{fig:ensemble}
\end{figure}

In addition, we perform experiments with various number of models used for ensemble when creating pseudo-labels for the unlabeled set $\mathcal{U}$. 
Figure~\ref{fig:ensemble} shows the results for $R=1,2,3,4,5$ across all values of $k$.
$R=3$ has the best overall performance and a moderate computational cost.
Therefore, we use the ensemble of the top three models to create reliable pseudo-labels.
\begin{table*}[!t]
	\centering
	\caption
		{
			Classification accuracy (\%) of the baselines and our method on the test split of UCF101.
			We report the mean accuracy of each method across two runs with different randomly sampled $\mathcal{V}$.
		}
	\label{tbl:result_ucf}

	\begin{tabular}{l|l|c|c|c|c|c|c} 
		\toprule	
		\multirow{2}{*}{\textbf{Method}}\hspace{22ex}
		 & \multirow{2}{*}{\textbf{Datasets}}\hspace{4ex}
		 & \multicolumn{3}{c|}{per-frame}&
		 \multicolumn{3}{c}{per-video}\\
		 \cmidrule{3-8}
		 & & \textbf{k=3} & \textbf{k=5} & \textbf{k=10} & \textbf{k=3} & \textbf{k=5} & \textbf{k=10} \\
		\midrule
		Target only & $\mathcal{V}$ & 42.58 & 53.31 &63.05 & 43.74 &55.50 &64.74\\
		Source only & $S_1,S_2$ & 41.96& 41.96 & 41.96& 43.46 & 43.46 & 43.46\\		
		Fine-tune & $S_1,S_2,\mathcal{V}$ & 55.86&60.55 &66.77 &58.57 &66.01 &70.21 \\
		EnergyNet~\cite{Junnan_MM_2017} & $S_1,S_2,\mathcal{V}$ &  55.93 &60.82  &66.73 &  58.70 &66.23 &70.25\\
		GradMix & $S_1,S_2,\mathcal{V}$ & \textbf{56.25}  & \textbf{61.73} & \textbf{67.30} & \textbf{59.41} & \textbf{66.27} & \textbf{71.49}\\
		\midrule
		MDDA~\cite{Mancini_CVPR_2018} & $S_1,S_2,\mathcal{V},\mathcal{U}$ &56.65 & 61.58 &67.65  & 60.00&65.14 &71.54\\			
		DCTN~\cite{Xu_2018_CVPR} &$S_1,S_2,\mathcal{V},\mathcal{U}$ &57.88 & 61.97 &68.46  & 61.64&66.59 &72.85\\
		GradMix w/ hard label& $S_1,S_2,\mathcal{V},\mathcal{U}$  & 68.92 & 68.76 & 69.25 & 72.58 & 72.34& 73.48\\	
		GradMix w/ VAT~\cite{vat}& $S_1,S_2,\mathcal{V},\mathcal{U}$  & 69.02 & 69.59 & \textbf{70.11} & 73.35 & 73.05& \textbf{73.71}\\		
		GradMix w/ MixMatch~\cite{mixmatch}& $S_1,S_2,\mathcal{V},\mathcal{U}$  & \textbf{69.33} &  \textbf{69.88} & 70.09 & \textbf{73.57}& \textbf{73.46}& 73.68\\		
		
%

		\bottomrule
	
	\end{tabular}
	
\end{table*}	

\subsection{MPII + BU101 $\rightarrow$ UCF101}
\noindent\textbf{Experimental setting.}
In the action recognition experiment, 
we have four sets of training data similar to the digit recognition experiment, which include
(1) $S_1$: labeled images from the training split of MPII,
(2) $S_2$: labeled images from the training split of BU101,
(3) $\mathcal{V}$: $k$ labeled video clips per class randomly sampled from the training split of UCF101,
(4) $\mathcal{U}$: unlabeled images from the rest of the training split of UCF101.
We experiment with $k=3,5,10$ which corresponds to $303,505,1010$ video clips.
Each experiment is run two times with different $\mathcal{V}$.
We report the mean accuracy across the two runs for both per-frame classification and per-video classification.
Per-frame classification is the same as doing individual image classification for every frame in the video,
and per-video classification is done by averaging the softmax score for all the frames in a video as the video's score.

\noindent\textbf{Baselines.}
We compare our method with multiple baselines described in Section~\ref{sec:exp_digit},
including \textit{Target only},  \textit{Source only}, \textit{Fine-tune}, \textit{MDDA}~\cite{Mancini_CVPR_2018} and \textit{DCTN}~\cite{Xu_2018_CVPR}.
In addition,
we evaluate another baseline for knowledge transfer in action recognition, namely EnergyNet~\cite{Junnan_MM_2017}: The ConvNet (ResNet-18) is first trained on MPII and BU101,
then knowledge is transfered to UCF101 through spatial attention maps using a Siamese Energy Network.

\noindent\textbf{Results.}
Table~\ref{tbl:result_ucf} shows the results for action recognition.
\textit{Target only} has better performance compared to \textit{Source only} even for $k=3$,
which indicates a strong distribution shift between source data and target data for actions in the wild.
For all values of $k$,
the proposed GradMix outperforms baseline methods that use $S_1,S_2$ and $\mathcal{V}$ for training
in both per-frame and per-video accuracy.
GradMix also has comparable performance with MDDA that uses the unlabeled dataset $\mathcal{U}$.
The proposed pseudo-label method achieves significant gain in accuracy by assigning hard labels to $\mathcal{U}$ and learn target-discriminative knowledge from the pseudo-labeled dataset. 
Futhermore,
performance improved is achieved by incorporating state-of-the-art semi-supervised learning methods.
\section{Conclusion}
\label{sec:conclusion}

In this work,
we propose GradMix,
a method for semi-supervised MS-DTT: multi-source domain and task transfer.
GradMix assigns layer-wise weights to the gradients calculated from each source objective,
in a way such that the combined gradient can optimize the target objective, measured by the loss on a small validation set. 
GradMix can adaptively adjust the learning rate for each mini-batch based on its importance to the target task.
In addition, we assign pseudo-labels to the unlabeled samples using model ensembles, and consider the pseudo-labeled dataset as a source during training. 
We validate the effectiveness our method with extensive experiments on two MS-DTT settings, namely digit recognition and action recognition.
GradMix is a generic framework applicable to any models trained with gradient descent. 
For future work, we intend to extend GradMix to other problems where labeled data for the target task is expensive to acquire, such as image captioning.

\vspace{5ex}
\section*{Acknowledgment}
\label{sec:acknowledgement}
This research is supported by the National Research Foundation, Prime Minister's Office, Singapore under its Strategic Capability Research Centres Funding Initiative.
The computational work for this article was partially performed on resources of the National Supercomputing Centre, Singapore (https://www.nscc.sg).

{\small
	\balance
	\bibliographystyle{ieee}
	\bibliography{reference}
}

\end{document}